\documentclass{article}

\usepackage{arxiv}

\usepackage[utf8]{inputenc} 
\usepackage[T1]{fontenc}    
\usepackage{hyperref}       
\usepackage{url}            
\usepackage{booktabs}       
\usepackage{amsfonts}       
\usepackage{nicefrac}       
\usepackage{microtype}      
\usepackage{lipsum}
\usepackage{graphicx}
\usepackage{subfigure}
\usepackage{amsmath}
\usepackage{multicol}
\usepackage{multirow}
\usepackage[capitalize]{cleveref}
\graphicspath{ {./} }

\title{Q-Learning Based System for Path Planning with UAV Swarms in Obstacle Environments}

\author{
  Alejandro Puente-Castro\thanks{Corresponding author: a.puentec@udc.es} \\
  Faculty of Computer Science, CITIC\\
  University of A Coruna\\
  A Coruna, 15007, Spain\\
   \And
Daniel Rivero\\
  Faculty of Computer Science, CITIC\\
  University of A Coruna\\
  A Coruna, 15007, Spain
   \And
Eurico Pedrosa\\
  IEETA, DESI, LASI,\\
  University of Aveiro,\\
  Portugal
   \And
Artur Pereira\\
  IEETA, DESI, LASI,\\
  University of Aveiro,\\
  Portugal
   \And
Nuno Lau\\
  IEETA, DESI, LASI,\\
  University of Aveiro,\\
  Portugal
   \And
Enrique Fernandez-Blanco\\
  Faculty of Computer Science, CITIC\\
  University of A Coruna\\
  A Coruna, 15007, Spain
}

\begin{document}
\maketitle
\begin{abstract}
Path Planning methods for autonomous control of Unmanned Aerial Vehicle (UAV) swarms are on the rise because of all the advantages they bring. There are more and more scenarios where autonomous control of multiple UAVs is required. Most of these scenarios present a large number of  obstacles, such as power lines or trees. If all UAVs can be operated autonomously, personnel expenses can be decreased. In addition, if their flight paths are optimal, energy consumption is reduced. This ensures that more battery time is left for other operations. In this paper, a Reinforcement Learning based system is proposed for solving this problem in environments with obstacles by making use of Q-Learning. This method allows a model, in this particular case an Artificial Neural Network, to self-adjust by learning from its mistakes and achievements. Regardless of the size of the map or the number of UAVs in the swarm, the goal of these paths is to ensure complete coverage of an area with fixed obstacles for tasks, like field prospecting. Setting goals or having any prior information aside from the provided map is not required. For experimentation, five maps of different sizes with different obstacles were used. The experiments were performed with different number of UAVs. For the calculation of the results, the number of actions taken by all UAVs to complete the task in each experiment is taken into account. The lower the number of actions, the shorter the path and the lower the energy consumption. The results are satisfactory, showing that the system obtains solutions in fewer movements the more UAVs there are. For a better presentation, these results have been compared to another state-of-the-art approach.
\end{abstract}


\section{Introduction}\label{introduction}
New uses for swarms of unmanned aerial vehicles (UAVs) are being developed to solve different industrial and emergency problems \cite{albani2017monitoring,huuskonen2018soil, CORTE2020105815, bocchino2018f, rabinovitch2021scaling, liu2018motif}. The advantages provided by UAVs such as their low cost, excellent mobility, safety, and convenient size for some maneuvers, are the main cause of their growing popularity \cite{yeaman1998virtual}. All these advantages are provided by the wide variety of UAVs that exist for every need. That variety allows mounting different types of sensors with different capabilities. The development of sensors for UAVs is growing, especially remote sensing \cite{noor2018remote}. The flexibility in the characteristics of UAVs, such as their architecture or the sensors they allow, make them popular tools for different needs \cite{austin2011unmanned}.

However, UAVs have drawbacks, the most important being power consumption, which reduces operation time. Due to their small size, it is difficult to obtain small power sources that have large capacities while maintaining small weights. When weights are small, flight operations are not compromised by having more flight time available.

The constraints on flight time imposed by the batteries may be lessened when used in groups or swarms. In other words, since flight paths are shorter when several UAVs operate simultaneously, numerous jobs can be completed quicker. By doing this, the probability that the UAVs' batteries will not be enough to propel them over the terrain is decreased. A UAV is less likely to stop in the middle of an operation as a result of decreased energy availability, which causes fewer risks of falling. 

Like any type of robotic swarm, UAV swarms may be used in the real world for any activity, as in individual use. The main benefit of the swarm robotics technique is robustness, which takes many different forms. First off, a swarm can self-organize or dynamically reorganize how individual robots are deployed since it is composed of many relatively simple agents that are not pre-assigned to certain roles or duties. Additionally, and for the same reasons, the swarm technique is very tolerant of individual agent failure. There is no common-mode failure point or vulnerability in the swarm because control is entirely decentralized. In contrast to the significant technical expense of fault tolerance in conventional robotic systems, it might be argued that the high level of robustness seen in UAV swarms is inherent to the swarm robotics methodology \cite{sahin2008special}.  

The number of UAV operators needed for the initial flight tests with swarms was equal to the number of UAVs, which greatly raised the cost of operation when used in groups. There have been recent advancements in the creation of algorithms \cite{zhao2018survey} and communications \cite{campion2018review} that allow for the control of the entire swarm with just one person capable of operating the systems. These developments enable more effective and quicker communication between UAVs as well as better collision avoidance path calculation, reducing the need for human involvement in risk situations. Thus, the most recent approaches already aim at autonomous control of the entire swarm. These paths have to be calculated at the lowest possible cost and are intended to be as efficient as possible. This is known as the Path Planning Problem \cite{aggarwal2020path}. In which the aim is to plan the path of movement of robotic systems such as UAVs. Due to the low altitude of many of their operations, the aircrafts have to deal with obstacles present in the flight area. Therefore, the calculation of flight paths must be taken into consideration for the presence of these obstacles and the future positions of all UAVs in the swarm, to avoid collisions between members of the fleet. Paths must be as optimal as possible while avoiding obstacles and other UAVs.

To deal with the complexity of this kind of development, different algorithms are offered in the Swarm Intelligence (SI) \cite{kennedy2006swarm}. The aim of these algorithms is to simultaneously coordinate a large number of agents. This coordination is based on a group of individual actors in a self-organized and strong manner while keeping to basic, common rules \cite{bonabeau2001swarm}. That is, each UAV in the swarm is an individual actor.  Each actor has its own information and its behavior depends on its information, the rules of the system, and the information shared by the others. This coordinated behavior is intended to achieve an objective in the most optimal way \cite{stentz1997optimal}.

Some of these Path Planning algorithms are being used in military applications. The limited civilian applications are mainly to pursue or attain goals, such as mapping trails across cities \cite{puente2021review}. Despite the wide range of possibilities, there are not many technologies designed specifically for agricultural and forestry applications that are focused on optimizing field prospecting tasks. 



The goal of this research is to develop a system for solving the Path Planning problem in 2D grid-based maps with fixed obstacles and variable numbers of UAVs using Q-Learning techniques supported by Artificial Neural Networks (ANN). Therefore, the main contributions of this work can be enumerated as:

\begin{enumerate}
    \item An innovative Q-Learning-based system that can determine the most optimal flying path for a UAV swarm to cover as much area as possible during prospecting activities.
    \item A system that can estimate the flight path of any number of UAVs on any size map with varied obstacle sets.
    \item A comparison of the results of using a single ANN for each UAV against a global ANN for all UAVs.
\end{enumerate}

The structure of this paper is as follows: An overview of the state of the art is provided in Section~\ref{background}; a description of the technical aspects required for the development of the proposed method is provided in Section~\ref{materials}; a summary of the results of the experimental process is provided in Section~\ref{results}; the conclusions drawn after evaluating the results are provided in Section~\ref{conclusions}; and, finally, the possible works and studies from which the problem to be addressed can be derived are listed in Section~\ref{future_work}.


\section{Background}\label{background}


In the state of the art, a large number of works can be found that are grouped into two branches of research \cite{puente2021review}. Of these two branches, Reinforcement Learning (RL) \cite{sutton2018reinforcement, wiering2012reinforcement} stands out. An example can be seen in the research of Qiu et al. \cite{qiu2022data} where they employ an Actor-Critic Reinforcement Learning algorithm for simultaneous control of multiple UAVs. In this paper, each UAV only has local information about the environment. It means, that the UAV only stores its own information and does not share anything with the rest of the group. So, some information about the environment may be lost or it may be captured redundantly. Wei et al. also make use of this Actor-Critic RL for collaborative data collection over large areas \cite{wei2022high}. They propose a way to calculate energy consumption that is only based on time. The drawback is that they do not consider the type of movement, so flights with more changes in direction may affect consumption more than flights in a straight line. For avoiding the sparse reward problem, common in this type of problem, they employ an incentive mechanism. Incentives are also applied by Salimi et al. for the control of a type of UAV group called flocks instead of swarms \cite{salimi2021deep}. In their paper, they state that they have used environments with up to 50 obstacles but do not show examples. Also, it seems that they rely on the evolution of the reward to know if the system is working correctly, but this does not guarantee that the objectives are completed in an optimal way, as it may get stuck in a local optimum. It is necessary to have more information on the results obtained at the global level of the system.

The most popular algorithm within RL is Q-Learning \cite{watkins1992q}.  It is the most common technique of all and many authors make use of it. Most of the papers use a variant known as Deep Q-Learning (DQN), which makes use of Artificial Neural Networks (ANN) to have more generalization capabilities in the control of multiple UAVs like in the work of Raja et al. \cite{raja2019inter}. Despite not presenting the findings, their paper claims that their technology is scalable to 100 UAVs. In addition to this, a generic system with significant commercial potential can be created by having a system that is scalable to any number of UAVs. Puente et al. \cite{puente2022uav} propose a dense two-layer ANN applied to Q-Learning. The proposed model is tested exclusively on maps without obstacles, so it may encounter major limitations on maps with obstacles. If the correct operation is demonstrated, it is only time-dependent, so the solution would depend on the hardware of the required equipment. Thus, equipment with better capabilities would have better times, which implies high costs.

On the other hand, the other branch is Evolutionary Computing (EC) \cite{davis1991handbook, jh1975adaptation}. For example, Kong et al. propose a Genetic Algorithm (GA) for swarm control in 3D environments \cite{kong2022improved}. This algorithm was tested in a simulator. They indicate that their system can avoid converging to local maxima. Despite this, it may have a high computational cost compared to other methods. Liu proposes another GA for 3D environments with terrain obstacles \cite{liu2022improved}. His method is able to obtain smoothed paths without having a subsequent smoothing phase. GA can also be used to complement other algorithms. In their paper, they propose a system whose fitness function is the distance of the UAVs to the final target. This is a very simplistic metric, as it can present a very slow approach to the target if the UAVs move in a spiral. Thus, a very high battery consumption is present while the UAVs are approaching slowly. In general, the flight environment strongly influences the behavior of the algorithm. That is, a 3D map implies controlling the height of the aircraft while a 2D grid-map implies knowing the state of each cell. To conclude with these techniques, there is a branch within EC known as Swarm Intelligence (SI) \cite{kennedy2006swarm} that seeks to mimic the collective behavior of natural systems. For example, Xu et al. propose to use GA to optimize a system based on Wolf Pack Algorithm, a pure SI technique, to control multiple UAVs \cite{xu2022task}. The paper demonstrates the goodness and efficiency of their system compared to others. However, they do not show examples of the environments in which the systems were tested.

Among the pure SI techniques, there is also a great deal of diversity, although they are not as widely used as those mentioned above. For example, Yang et al. make use of Particle Swarm Optimization (PSO) combined with a voting mechanism for multi-UAV control \cite{yang2019collision}. They created a spatially refined voting system for the traditional PSO. Additionally, their approach takes time into account and successfully creates collision-free pathways for several UAVs in the same amount of time. Pamosoaji et al. also make use of this algorithm for UAV control. In it, they take into account the limitations of the slower aircraft to reduce its flight time \cite{pamosoaji2019pso}. In their paper, they show that the system is capable of obtaining paths, but they do not show a measure of satisfaction of these paths. Jain et al. propose to use the Multiverse Optimizer algorithm (MVO) for the control of multiple UAVs and contrast it with the use of a single UAV as well. This demonstrates the generalization power of their system \cite{jain2019mvo}. Despite the high capabilities of its system, the limitations of major environmental factors are not taken into account during air operations. In general, the major drawback of SI techniques is that they tend to converge to a local optimum \cite{yang2014swarm}. In addition, it is very difficult to describe the collective behavior of natural systems; it may not be realistic.

Several papers in the state-of-the-art present a subsequent path smoothing stage. By establishing this later step, sharp turns in the paths are modified to make them softer and more gradual. In this way, paths with fewer curves are obtained, which lengthens the battery time. However, it involves more processing and, if the original path had errors, they can be propagated. In the proposed system, the path smoothing stage is not performed in order to reduce the computation time and not to alter the subsequent data capture by flying over as much of the terrain as possible.

\section{Materials and Method}\label{materials}

For the development of the system, the spotlight has been put onto: first, introduce the problem formulation; second,  define the flight environments set; third, define the UAV movements; fourth, introduce the proposed method for path calculation; fifth, define the most optimal parameters to solve the problem; and, finally, to define mechanisms to verify the correctness of the proposed approach and the satisfaction criteria of the results obtained.

\subsection{Problem Formulation}

The main aim of this research is to develop a system capable of solving the Path Planning Problem for UAV swarms in maps with obstacles. Path Planning problems involving multiple vehicles, such as UAVs, must take into account a number of variables in order to maintain efficiency and effectiveness while upholding control, collaboration, and safety standards. As a result, it is crucial to be able to address the problems caused by these variables as they are an important component of the main goal.

This way of looking at a Path Planning problem as the union of different inherent problems is common in the literature \cite{he2021novel, puente2022uav}. Thus, the experimentation process is more precise and organized. The formulation of the Path Planning problem presented is divided into the following areas:

\begin{itemize}
    \item[] Flight Environments Set
    \item[] UAV movements
    \item[] Proposed Model Design
    \item[] Model Optimization
    \item[] Model Evaluation Metric
\end{itemize}


\subsection{Flight Environments Set}\label{maps}

Most of the studies in Section \ref{background} employ maps with fixed dimensions (5$\times$5, 10$\times$10, or 20$\times$20 cells), but some also use continuous maps without cell division. It has been preferred to employ continuous maps that are segmented into cells since this paper aims to fly over the entire maps for data collection. In other words, it is of great interest to delimit portions of the total area in order to be able to collect as much data as possible in an organized manner.

The chosen flight maps have fewer cells than previous works that were mentioned. One significant restriction is the cost of flying over vast maps. Because each cell requires one stop to capture the map's surface, the UAVs have to perform several stops while capturing very large maps, which significantly depletes their battery. If maps are divided into fewer cells, the number of stops and starts made by each UAV is decreased by dividing the map into fewer cells, which lowers energy consumption.

\begin{figure}[!ht]
        \centering
        \subfigure[5$\times$5]{
        \includegraphics[scale=0.4]{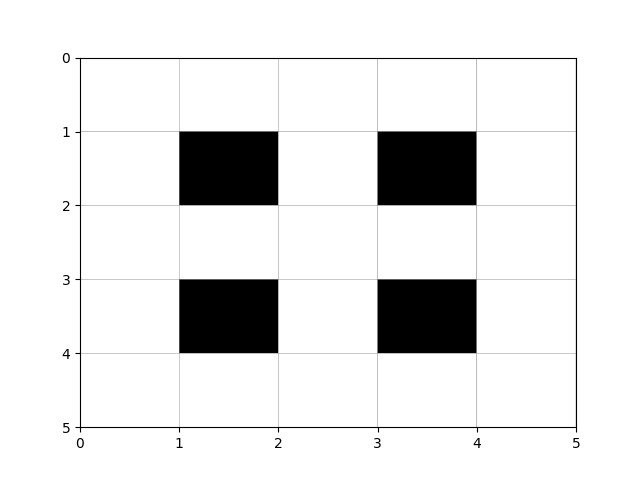}
        \label{fig:flight_environment1}
        }
    	\subfigure[6$\times$6]{\includegraphics[scale=0.4]{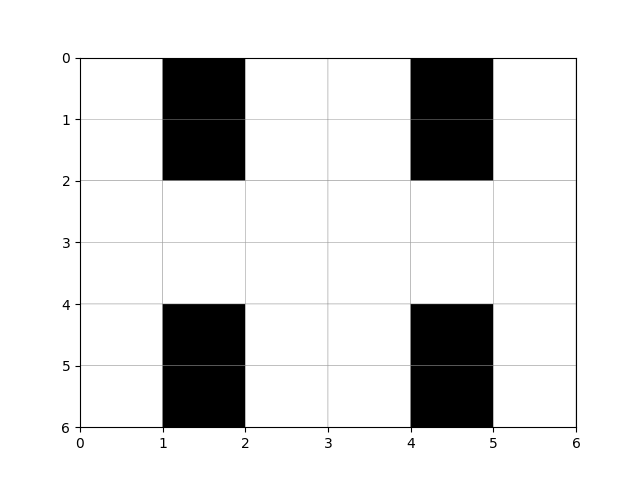}
        \label{fig:flight_environment2}}
    	\subfigure[7$\times$7]{\includegraphics[scale=0.38]{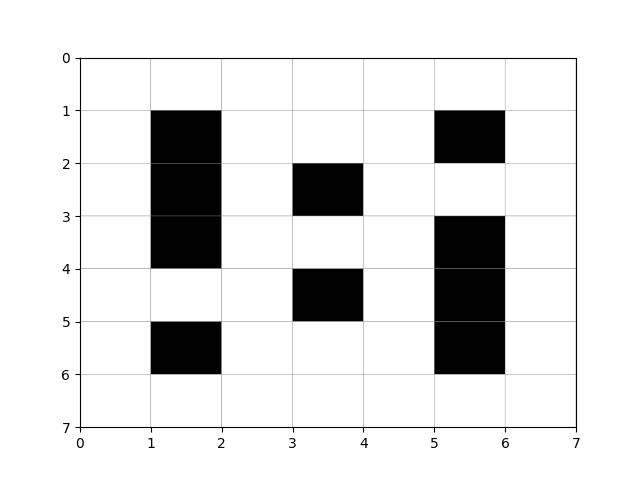}
        \label{fig:flight_environment3}}
    	\subfigure[8$\times$8]{\includegraphics[scale=0.38]{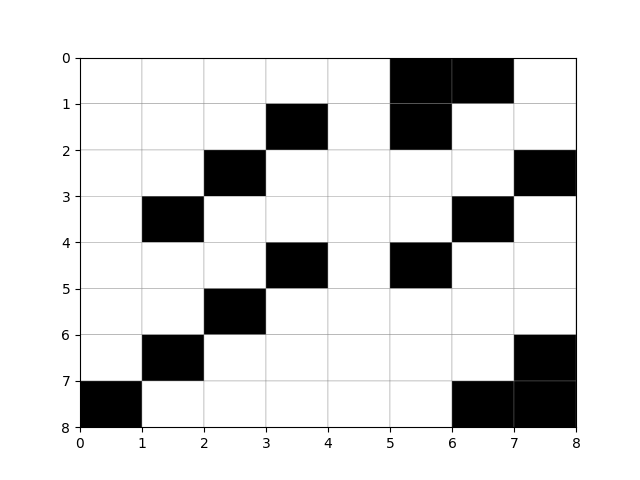}
        \label{fig:flight_environment4}}
    	\subfigure[9$\times$9]{\includegraphics[scale=0.38]{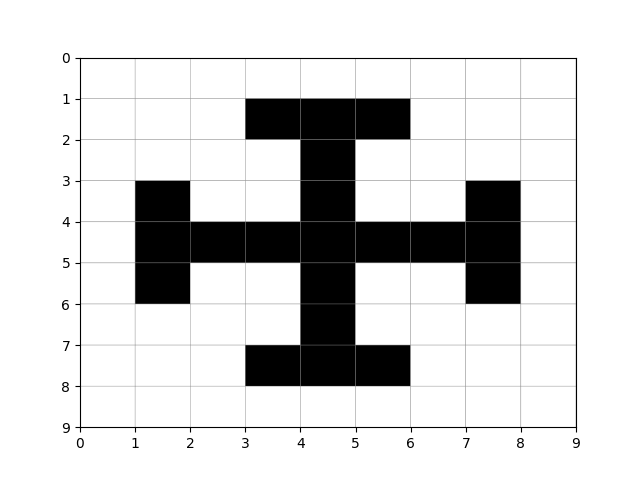}
        \label{fig:flight_environment5}}
        \caption{Maps used in the flight environments. Obstacles are shown in black. In white are the cells that can be flown over. UAVs must visit as many white cells as possible.}
    \label{fig:map_set}
\end{figure}

For this purpose, 5 maps have been defined, ranging in size from 5$\times$5 cells to 9$\times$9 cells (Figure~\ref{fig:map_set}). For the design of the obstacles, we have opted for configurations that force you to make many changes in direction and even go backward on the paths. This is because they force the UAV to take non-linear paths, which are the ones that have higher energy requirements.

In the case of the 5$\times$5 cell map (Figure \ref{fig:flight_environment1}), it is intended to simulate the case of tree crops such as olive trees, which are regularly arranged. To complicate that task, the 6$\times$6 cell map has been designed (Figure \ref{fig:flight_environment2}), which has situations that involve turning the UAVs around. In this way, the UAVs are forced to move backward. 

Both the 5$\times$5 and 6$\times$6 cell maps are horizontally and vertically symmetrical. To test how the UAVs behave outside these conditions, the 7$\times$7 cell map has been designed (Figure \ref{fig:flight_environment3}). Furthermore, following this premise, the 8$\times$8 cells map has been designed (Figure \ref{fig:flight_environment4}), which also tests the behavior of the system if the obstacles are arranged diagonally.

The last map to be tested is the 9$\times$9 cells map (Figure \ref{fig:flight_environment5}). In the previous maps, UAVs can pass through the gaps between the obstacles. This map tests the behavior of the system if a single large obstacle has to be circled. In addition, corners have been added to make it more difficult for UAVs to retrace their steps at some points.

Finally, despite varying the obstacles, the number of cells in the maps is also varied to test that the system works for any size. Therefore, the system is tested to prove that it is valid in different situations.

\subsection{Fundaments}

\subsubsection{Reinforcement Learning}

The solution for the Path Planning Problem for UAVs was Reinforcement Learning (RL) \cite{sutton2018reinforcement}. As in other computational techniques, it implies not having to define in the system the desired behavior. With this method, a component of the algorithm, also known as agent, learns the intended behavior based on a trial-and-error system of tests carried out in an interactive and dynamic environment \cite{wiering2012reinforcement, kaelbling1996reinforcement}. The agent must exploit what it already knows in order to profit from rewards, but it must also explore in order to choose its future actions more wisely. The problem is that pursuing either exploration or exploitation solely would result in failure. The agent must test several different options and gradually favor the ones that seem to work the best. For each action on a stochastic task to gain a valid estimate of the expected reward, several trials must be made. Finding the ideal behavior is complicated by the possibility that the agent's activities may have an impact on the surrounding environment.

The explicit consideration of the entire issue of a goal-directed agent interacting with an unpredictable environment is another important aspect of RL. Contrary to many techniques, RL does not take into account sub-problems without considering how they might relate to a bigger one. In other words, it addresses the problem "as a whole".

The learning method differs only slightly in most RL algorithms \cite{sutton2018reinforcement}. These strategies come in a variety of forms that let the systems handle a wide range of problems. It has been decided to employ a technique known as Q-Learning for this paper \cite{watkins1992q}. The biggest factor is that, in contrast to other variants, it does not require a model of the environment (model-free approach).

\subsubsection{Q-Learning}\label{qlearning}

The agents have to find and follow strategies that allow them to solve problems. These strategies are known as policies. The agents can use their experience to learn the values of all the policies in parallel even when they can only follow one policy at a time thanks to the traditional Q-Learning algorithms \cite{watkins1992q}. The agent learns to follow a policy only through trial and error in this model-free approach \cite{glascher2010states}. This greedy convergence of Q-Learning towards the best possible solution enables the solution to be reached independent of the decision-making policy, which is why it is known as an off-policy algorithm. In other words, it only bases its decisions on the agent's interactions with the environment around it. By doing this, it is ensured that the system will function in a variety of environments without the need to find the best policy to use in each one. The ``Q'' in Q-learning stands for quality and tries to represent how useful a given action is in gaining some future reward.

The most well-known benefit of Q-Learning over other RL techniques is that it allows for the comparison of predicted utility across different actions without the need for an environment model. That means the key factor that led to its selection for this study is how easily it learns and infers situations without requiring their modeling. These algorithms' primary distinction from other RL algorithms is that they choose the optimum course of action based on the values in a table. The values are referred to as Q-values and the table as Q-table. The values indicate how rewarding an action would be in the given environmental conditions. From these values, the action with the highest value for each state is chosen. Typically, Bellman's equation (Eq.~\ref{eq:qlearning}) is combined with the system's prior predictions to train it. The equation has different elements: $Q(s,a)$ is the function that calculates the Q-value for the current state ($s$), of the set of states $S$, and for the giving action ($a$), of the set of actions $A$, $r$ is the reward of the action taken in that state and it is computed by the reward function $R(s,a)$, $\gamma$ is the discount factor and $\arg\max_{a^{'}}(Q(s^{'},a^{'}))$ is the maximum computed Q-value of the pair $(s^{'},a^{'})$ represented as $Q(s^{'},a^{'})$. The pair $(s^{'},a^{'})$ is a potential next state-action pair. $(s^{'}$ is the next state and it is given by the transition function $T(s,a)$ which returns the state resulting from the execution of the selected action. The $a^{'}$, is each one of the available actions. 

\begin{equation}
\label{eq:qlearning}
Q(s,a) \leftarrow r + \gamma \times \arg\max_{a^{'}}(Q(s^{'},a^{'}))
\centering
\end{equation}

With probability $\epsilon$, a portion of the actions in a Q-Learning problem are made at random, and with probability $1-\epsilon$, the action with the greatest Q-value for that state is adopted. An episode is the series of actions that an agent performs for a certain $\epsilon$ until it achieves an end condition (task completion, end of time, etc.) \cite{shang2022hybrid}. The operation is started over at the beginning of each episode. During testing, episodes reduce the value of $\epsilon$ by a factor of reduction. In this approach, the decision of what to do is influenced more by the computed Q-values and less by chance. By considering the minimal value of $\epsilon$, it is kept from becoming too close to zero and avoid overfitting \cite{zhang2018study}. 

The key to convergence in Q-Learning is that it is a variant of a Markov Decision Process (MDP). This process is artificially controlled and is known as action-replay process (ARP) \cite{watkins1992q}. It should be noted that this description assumes a representation of a look-up table, indicating that Q-Learning might not converge correctly for other representations. The requirement that there must include an unlimited number of episodes for each beginning state and action is the most significant implicit condition in the convergence.

Recently, a variant known as Deep Q-learning (DQN) has emerged as an alternative. This approach varies from traditional Q-Learning in that it aims to enhance the calculation of the Q table using Machine Learning \cite{michie1994machine} or Deep Learning \cite{lecun2015deep}. The model may deduce the values of the Q table by abstracting sufficient knowledge. In some cases, Bellman's Equation bias concerns can be resolved in this way \cite{fan2020theoretical}.

\subsubsection{Artificial Neural Network}

One of the key points of this paper is the use of ANNs to enhance Q-learning \cite{krogh2008artificial}. The authors of this work choose a two-layer fully connected ANN. Unlike convolutional deep ANNs \cite{albawi2017understanding} that other authors have suggested in their studies, it is not assumed that the neighborhood of a cell provides additional information. Hence, it does justify the need to use dense layers \cite{huang2017densely}. The only input is the combination of the original environment map, the map with the position of each UAV, and the map with the visited cells and the output are the Q-values for all possible movements. As a result, each Q-Learning experiment follows:

\begin{enumerate}
    \item Build the ANN model(s) using the selected configuration.
    \item Determine the Q-table values and the optimum course of action for each UAV in the swarm using ANN model(s).
    \item Train the model(s) based on each chosen action's outcomes.
    \item Select the cases where the number of movements required to explore the entire map is lower.
\end{enumerate}

The information available to each agent or UAV in the swarm is very important. If the information is only local (the one perceived by the UAV itself), it implies the loss of information from other UAVs, which can be very useful. Contrastingly, if it is global (it knows the information available to all the UAVs in the swarm) it needs to establish mechanisms of communication of the information to perform a quick update of the knowledge. In this way, errors in path planning are avoided. According to previous studies on the state of the art, the system might be employed in two different ways without clear benefits for none of them. The first step is to create a single ANN that will be used to control all of the UAVs moving, determining the movement of each one in each time and verifying the reward received. As a result, the design and weights of all UAVs will be the same, and the behavior of each UAV will be determined by its current state. On the other hand, since each UAV can have a unique ANN, its reaction would depend on its design and weight as well as the current state. That is, the main objective of the experiments is to determine which ANN configuration is better as a controller with respect to the UAVs: one ANN for all UAVs (global ANN), or one ANN for each UAV (local ANN). In both cases, the input data is the same, the information obtained from all UAVs.

\subsubsection{Rewards}

The Reward Function is an incentive system in RL problems that instructs the agent through reward and punishment what is right and wrong. Agents seek to maximize overall gains, i.e. the summatory of all rewards in the episode, even at the expense of current actions.

The largest reward must be given in order for the UAV to move to previously unexplored locations. It is also crucial that it grows as fewer cells remain undiscovered (Eq.~\ref{eq:reward}). In other words, it follows a Hill-Climbing scheme \cite{kimura1995reinforcement}. For previously visited cells, another reward is needed. The UAV has a reward in the event that flying over a cell that has previously been visited in order to reach an unvisited one is preferable to flying around it (for example, when there are spurious cells left unvisited). They are given the lowest incentive to prevent UAVs from flying into cells that they are unable to visit. In these situations where the incentive is the lowest and the goal is to maximize rewards, UAVs learn that it is best to avoid these situations and opt for the ones that offer higher rewards, which will allow them to maximize the total reward outcome.

\begin{equation}
\text{new cell reward} = \text{new cell base reward} \times (1 + \frac{\text{\textit{max}(rows, columns)}}{\text{non visited cells}})
\label{eq:reward}
\end{equation}

\subsubsection{Memory Replay}

The Memory Replay technique is used in the majority of the state of the art to improve the experience that agents have gained from their environment. In the Memory Replay approach, the model is trained using a collection of previously recorded observations. A variety of information, including the activities conducted and their reward, is used in the observations. Regularly reusing experiences, increases sample efficiency and helps in stabilizing the model's training process \cite{foerster2017stabilising}. The memory should hold as many recent observations as possible, but in order to best utilize computational resources, it has a maximum size. Because of this, the memory uses a First-In-First-Out (FIFO) method to get rid of outdated observations with a maximum size of 60 elements.

Commonly in the state of the art, each UAV in the group has a separate memory when using the Memory Replay approach. It records observations together with the operations the UAV itself takes in its memory. The actions of other UAVs are never recorded. This keeps the information from becoming cluttered. Given that multiple UAVs may be located at different locations on the map, the fact that an action is erroneous for one UAV does not always mean that it is improper for others. Moreover, by combining the observations of all UAVs, one UAV may discover actions or combinations of actions that can serve the others later on. The end results might be significantly impacted by the memory's size and structure \cite{liu2018effects}. 

\subsubsection{Evaluation Metric}

To estimate the goodness of the proposed method, it has been decided to count the number of actions (also known as movements) performed by all UAVs in the system (eq. \ref{eq:total_actions}). The number of actions performed by a single UAV is the same as the length of its flight path (eq. \ref{eq:drone_actions}). For a flight path, having too many actions implies higher energy consumption and errors. For instance, it is less optimal than another path with fewer actions and that flies over the same cells. 

Authors in the state-of-the-art smooth the paths to make them simpler and more optimal \cite{correl2016introduction}. A grid-map will produce paths with several abrupt turns, but a sampling-based technique will produce paths that are randomly zigzagged. Running an additional algorithm that smooths the path and eliminates some of the turns can significantly enhance the results. On the other side, smoothing the paths may imply accuracy errors when retrieving data, since not all the cells being overflown may be totally covered.

\begin{equation}
    \text{drone$_i$ taken actions} = \text{lenght(drone$_i$ path)}
    \label{eq:drone_actions}
\end{equation}

\begin{equation}
    \text{Total actions} = \sum_{i=1}^{n} \text{drone$_i$ taken actions}
    \label{eq:total_actions}
\end{equation}

Because it is desired to lower the energy consumption for each operation in order to shorten the load time between processes. UAVs are considered to stop once the task is completed and are not considered to automatically return to the starting point. Therefore, the energy consumption of flying back is reduced.

\subsection{Proposed Method}

As a proposed model for the experiments, a system based on Q-Learning techniques that relies on ANN for better results has been chosen. The best ANN architecture and the best parameters for all precise aspects have been sought through random hyperparameter search \cite{bergstra2012random}. Thus, the most optimal possible combination of parameters to train the ANNs and have optimal results are obtained.

\begin{figure}[htp!]
    \centering
    \includegraphics[scale=0.13]{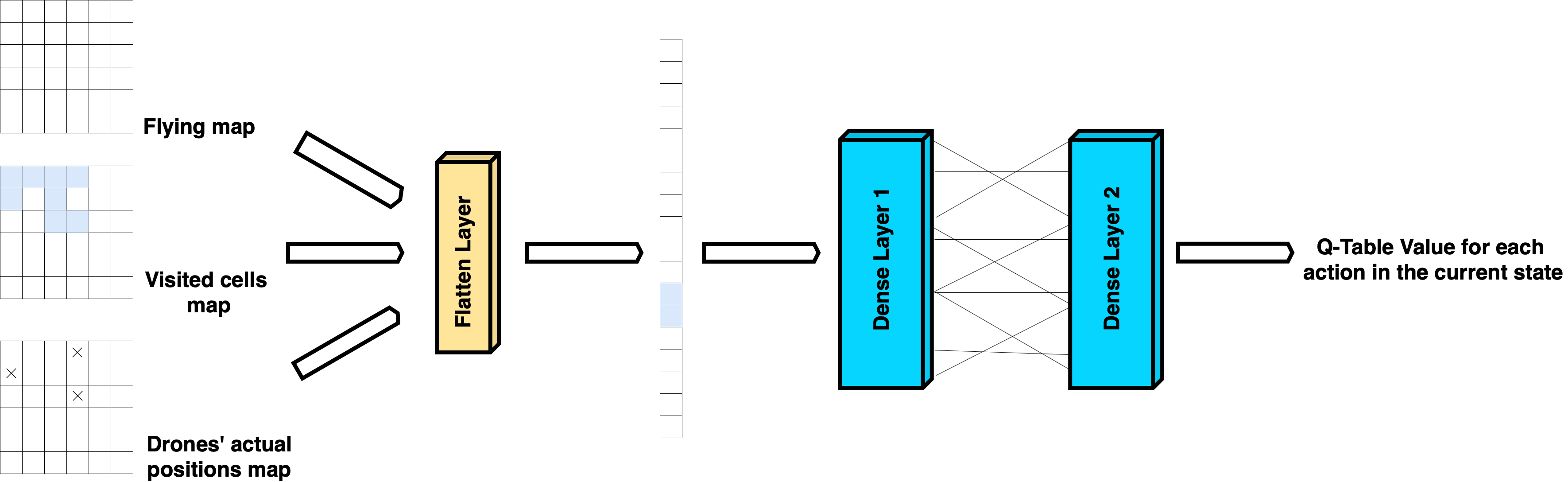}
    \caption{Diagram with the proposed ANN model. The three inputs are combined into one and the model calculates the Q-values corresponding to each action for the current state.}
    \label{fig:ann}
\end{figure}

An ANN made up of two dense layers \cite{huang2017densely, heaton2008introduction}, one with 1013 neurons and a relu activation function \cite{agarap2018deep}, and the other with 4 neurons and a softmax output function \cite{gao2017properties}, has been chosen through empirical experimentation (Figure \ref{fig:ann}). The Stochastic Gradient Descend (SGD) \cite{sutskever2013importance} optimizer was selected as the ANN's optimizer. Two hidden layers have been chosen because as universal solutions equivalent to a Turing machine, these kinds of networks can approximate any mapping regardless of the required accuracy, which means it is not necessary to use a path smoothing stage \cite{heaton2008introduction}. The network's input is the combination of the original environment map,  the map with the visited cells, and the map with the position of each UAV. As a result, ANN does not require any additional data beyond what is already in the environment. On the other side, the outputs of the ANN are the Q-values of each action for a given state, which are modulated by the softmax function. Therefore, there are four independent values for the respective Q-value for each movement: north, east, south, and west.

To meet all the requirements of the Q-Learning issues explained in Section \ref{qlearning}, an epsilon  value ($\epsilon$) of 0.49 has been selected as the probability of making actions at random. The factor of reduction for $\epsilon$ equals 0.93, in order not to decrease the value too much and the model continues to learn from the exploration. The minimal value for $\epsilon$ of 0.05. The chosen value for the discount factor ($\gamma$) is 0.83. All values were selected after previous exploratory research.

The reward values for the agents are in Table \ref{tab:rewards}. A scheme of positive reinforcement combined with negative reinforcement is followed. That is, discovering new cells is rewarded and the opposite is punished. Thus, the agents are encouraged to discover new cells, instead of revisiting the already-known ones. In addition, passing through an already visited cell is penalized less than passing through a forbidden area. The main reason behind this behavior is that it may be necessary to have paths that cross each other and that is not a mistake. If the rewards were equal, there is a risk that agents would retrace their steps as it is a reward maximization problem. Therefore, this situation is penalized in case it is not avoidable for the cases in which it is essential to cross paths.

\begin{table}[!ht]
\centering
    \begin{tabular}{c|c}
          & \textbf{Reward} \\\hline
         new cell base reward & 29.40 \\\hline
         visited cell reward & -31.66 \\\hline
         non-visitable cell & -45.44 \\\hline
    \end{tabular}
    \caption{Assigned rewards to the various cell types that each UAV visits. The initial rewards values were determined from a prior random exploration \cite{bergstra2012random} in which the most advantageous reward combinations were chosen.}
    \label{tab:rewards}
\end{table}

\begin{figure}[ht!]
    \centering
    \subfigure[Local Memory]{
        \includegraphics[scale=0.4]{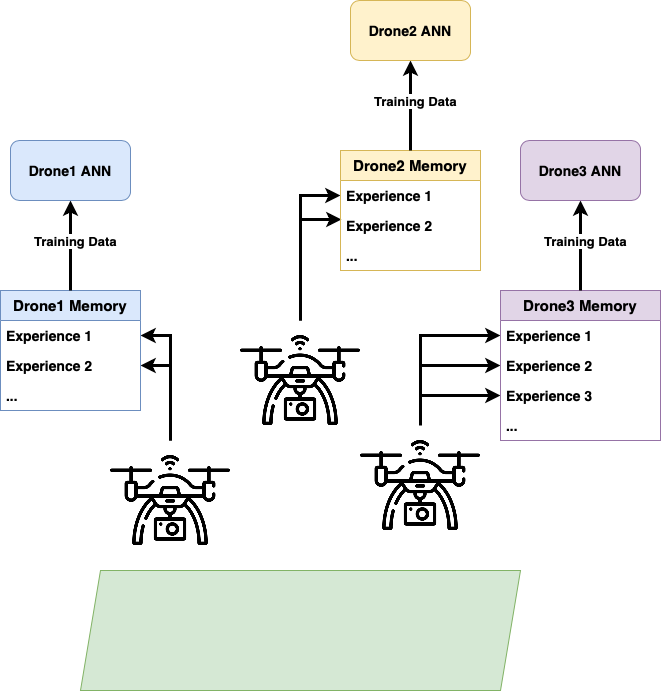}
        \label{fig:local_memory}
    }
    \subfigure[Global Memory]{
        \includegraphics[scale=0.4]{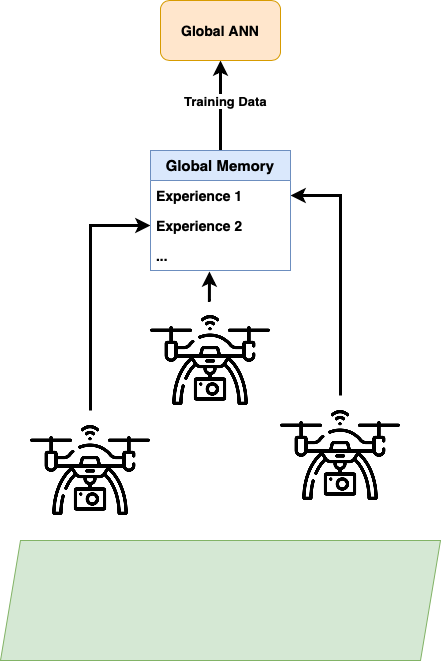}
        \label{fig:global_memory}
    }
    \caption{Diagram illustrating the differences in the UAV experience memory system: Fig. \ref{fig:local_memory} shows how UAVs write their experiences in their own memories, which will be used to train their own ANNs, so each memory only has experiences from one UAV. Fig. \ref{fig:global_memory} illustrates how the UAVs record their experiences in order in a single memory, which will then be used to train a neural network, mixing the experiences of all UAVs together.}
    \label{fig:experience_replay}
\end{figure}

A memory size of 60 actions with their corresponding rewards was selected for this investigation. Because UAVs frequently make mistakes in the initial stages of the process, it is crucial to have a high value for storing a number of experiences according to the number of map cells. Therefore, avoiding them and arriving at a more effective solution will be made possible by learning from the majority of errors and again training the model with them. Although the memory has a maximum size of 60 elements, it must be possible to determine how it behaves with respect to the ANN (Fig. \ref{fig:experience_replay}). Therefore, if it is an ANN per UAV, each ANN will have its memory with the unique experience of a single UAV (Fig. \ref{fig:local_memory}). Contrarily, when dealing with a single ANN for all UAVs, it has been decided to use a single collective memory (Fig. \ref{fig:global_memory}). Thus, the network learns the cases faced by all UAVs, and, in addition, the data are arbitrarily arranged, similar to having a random buffer in classical Memory Replay \cite{liu2018effects}. By having the elements arranged randomly, the model is prevented from memorizing movement patterns and learning to generalize flight behavior. In this case, the elements are random but you have elements that represent the experience of each UAV, not just the shuffled experiences of a single UAV.

In addition to the above, the case that the system does not find solutions for given conditions has been taken into account. Therefore, as a limiting condition for shutdown, the maximum flight time has been set at 30 minutes. This choice has been made on the basis of most commercial UAVs. These have flight autonomies of about 30 minutes, so it is assumed that this is the maximum time that UAVs in the swarm are capable of staying in the air.

\section{Results}\label{results}

For conducting the tests and analyzing the results, a set of combinations of map sizes with obstacles (Subsection~\ref{maps}) and UAV count have been defined. The number of actions carried out by each UAV was taken into consideration when analyzing the results.

\subsection{Experiment Design}

Twenty-five experiments have been created to evaluate the system's capabilities, as presented in this paper. The number of UAVs, the number of ANNs, and the size of the map vary between each one of them.

Since these are ANNs with random initialization, different seeds are tested to have a higher generalization power \cite{zhang2018dissection}. In addition, for better statistical measurement. The experiments are repeated 5 times with different seeds to have their mean and standard deviation.

All of the maps selected are grid maps, the same as in the studies cited in Section \ref{background}. The studies were carried out using grid maps of 5$\times$5, 6$\times$6, 7$\times$7, 8$\times$8, and 9$\times$9 cells (Subsection \ref{maps}). Different map sizes provide a better understanding of the system's capabilities while dealing with variable map sizes.

Each chosen map type was evaluated with an increasing number of UAVs because it is crucial for the system to function with any quantity of UAVs. Separate tests with the 1, 2, and 3 UAVs have been carried out. Thus, it is demonstrated that the system can adapt to a variety of UAV numbers. It is also worth highlighting the equivalence of employing a global or local approach when a single UAV is used. So, those executions have been referred to as baseline. As a result, it is assumed that the experiment will begin by controlling a single UAV, which is the simplest situation.

Additionally, there are fewer cells in the selected flight environment than in the studies that were mentioned. One significant restriction is the cost of flying over vast maps. Because each cell requires one stop to image the map's surface, the UAVs must make several stops while photographing very large maps, which significantly depletes their battery. The number of stops and starts made by each UAV is decreased by subdividing the map into fewer cells, which lowers energy consumption but increases the portion of land each cell represents. Each captured image is better when it captures larger portions of the map, as it contains more context information and is more suitable for further processing. That is, despite having less information detail, it has more information about the surrounding context, thus providing better information for combining the images and for processing them. It is important to adjust the number of cells to the size of the map. Otherwise, a single cell representing a very large area containing a very small obstacle means that information is lost because it is not overflown, as it would be considered as an obstacle cell.

To simplify processing, it has been decided to establish atomic movements (North, South, East, and West). The flight control of the UAVs themselves allows such well-defined actions without the need to turn to the sides. In addition, the existence of many curves further increases the energy requirements.

The range of possible movements or actions ($a$) that UAVs are capable of taking was codified as is common in this type of technique. That is, integer values from 0 to 3 are assigned to the movements: north, east, south, and west. This results in the encoding of every movement to a discrete list of fixed and different values.

All these variables are summarized in Table \label{ref:parameters}.

\begin{table}[htp!]
    \centering
    \begin{tabular}{|c|c|}
        \hline
        \textbf{Variable}                     & \textbf{Value}        \\ \hline
        Neurons First Dense Layer   & 1013 neurons \\ \hline
        Activation Function First Dense Layer   & Relu         \\ \hline
        Neurons Second Dense Layer  & 4 neurons    \\ \hline
        Activation Function Second Dense Layer  & Linear      \\ \hline
        ANN Output Function  & Softmax      \\ \hline
        Epsilon ($\epsilon$)         & 0.49         \\ \hline
        Epsilon decay                & 0.93         \\ \hline
        Minimum Epsilon ($\epsilon$) & 0.05         \\ \hline
        Discount Factor ($\gamma$)                     & 0.83         \\ \hline
        Memory Size                  & 60 actions   \\ \hline
        Maximum Flight Time          & 30 minutes \\ \hline
        Maximum Number of Episodes   & 30 episodes \\ \hline
        Possible actions   & North, East, South, West \\ \hline
    \end{tabular}
    \caption{Summary table with the values chosen for experimentation. All the values have been obtained through a preliminary testing process.}
    \label{tab:parameters}
\end{table}

\subsection{Experimental Results}

Table \ref{tab:table_results} shows the results obtained from the experimentation. For each map size, the mean and standard deviation of actions taken for each ANN configuration when faced with different numbers of UAVs are compared. To better show the capabilities of the proposed model (known as Proposed in Table \ref{tab:table_results}) it is compared with the model proposed by Puente et al.\cite{puente2022uav} which already demonstrated its capabilities on obstacle-free maps. It can be seen that the means of the results of the proposed model are lower than those of the model with which they are contrasted. This can be interpreted as an indication that the paths take fewer actions to complete the operation. Therefore, they are more optimal and efficient.

\begin{table}[!ht]
    \centering
    \begin{tabular}{cccccc}
    \cline{3-6}
                       &  & \multicolumn{4}{|c|}{
        ANN Configuration} \\ \hline
        \multicolumn{1}{|c|}{Map Size} & \multicolumn{1}{|c|}{Number of UAVs} & \multicolumn{1}{|c|}{Local \cite{puente2022uav}} & \multicolumn{1}{|c|}{Global \cite{puente2022uav}} & \multicolumn{1}{|c|}{\textbf{Local Proposed}} & \multicolumn{1}{|c|}{\textbf{Global Proposed}} \\ \hline
        \multicolumn{1}{|c|}{\multirow{3}{*}{5$\times$5}} & \multicolumn{1}{|c|}{Baseline} & \multicolumn{2}{|c|}{283.20 ± 97.79}  & \multicolumn{2}{|c|}{189.00 ± 91.06} \\ \cline{2-6} 
        \multicolumn{1}{|c|}{\multirow{3}{*}{}}  & \multicolumn{1}{|c|}{2 UAVs} & \multicolumn{1}{|c|}{147.60 ± 38.68} & \multicolumn{1}{|c|}{153.80 ± 56.42} & \multicolumn{1}{|c|}{79.40 ± 7.82} & \multicolumn{1}{|c|}{86.60 ± 34.62} \\ \cline{2-6} 
        \multicolumn{1}{|c|}{\multirow{3}{*}{}}  & \multicolumn{1}{|c|}{3 UAVs} & \multicolumn{1}{|c|}{76.60 ± 53.26} & \multicolumn{1}{|c|}{100.60 ± 51.76} & \multicolumn{1}{|c|}{56.20 ± 21.32} & \multicolumn{1}{|c|}{61.60 ± 47.26} \\ \hline
        \multicolumn{1}{|c|}{\multirow{3}{*}{6$\times$6}} & \multicolumn{1}{|c|}{Baseline} & \multicolumn{2}{|c|}{503.60 ± 195.34} & \multicolumn{2}{|c|}{212.60 ± 49.42}\\ \cline{2-6}
        \multicolumn{1}{|c|}{\multirow{3}{*}{}}  & \multicolumn{1}{|c|}{2 UAVs} & \multicolumn{1}{|c|}{145.40 ± 24.93} & \multicolumn{1}{|c|}{232.60 ± 189.49} & \multicolumn{1}{|c|}{123.00 ± 12.98} & \multicolumn{1}{|c|}{230.00 ± 156.62} \\ \cline{2-6}
        \multicolumn{1}{|c|}{\multirow{3}{*}{}}  & \multicolumn{1}{|c|}{3 UAVs} & \multicolumn{1}{|c|}{122.00 ± 55.29} & \multicolumn{1}{|c|}{139.60 ± 51.71} & \multicolumn{1}{|c|}{127.20 ± 69.83} & \multicolumn{1}{|c|}{135.60 ± 51.08} \\ \hline
        \multicolumn{1}{|c|}{\multirow{3}{*}{7$\times$7}} & \multicolumn{1}{|c|}{Baseline} & \multicolumn{2}{|c|}{523.60 ± 127.93} & \multicolumn{2}{|c|}{491.20 ± 15.61}\\ \cline{2-6}
        \multicolumn{1}{|c|}{\multirow{3}{*}{}}  & \multicolumn{1}{|c|}{2 UAVs} & \multicolumn{1}{|c|}{384.80 ± 112.3} & \multicolumn{1}{|c|}{537.40 ± 425.41} & \multicolumn{1}{|c|}{348.80 ± 151.46} & \multicolumn{1}{|c|}{278.40 ± 143.34} \\ \cline{2-6}
        \multicolumn{1}{|c|}{\multirow{3}{*}{}}  & \multicolumn{1}{|c|}{3 UAVs} & \multicolumn{1}{|c|}{199.40 ± 66.09} & \multicolumn{1}{|c|}{292.20 ± 181.60} & \multicolumn{1}{|c|}{166.60 ± 56.00} & \multicolumn{1}{|c|}{151.20 ± 72.70} \\ \hline
        \multicolumn{1}{|c|}{\multirow{3}{*}{8$\times$8}} & \multicolumn{1}{|c|}{Baseline} & \multicolumn{2}{|c|}{1011.00 ± 258.16} & \multicolumn{2}{|c|}{1367.80 ± 543.17}\\ \cline{2-6}
        \multicolumn{1}{|c|}{\multirow{3}{*}{}}  & \multicolumn{1}{|c|}{2 UAVs} & \multicolumn{1}{|c|}{700.00 ± 221.08} & \multicolumn{1}{|c|}{611.80 ± 484.85} & \multicolumn{1}{|c|}{757.60 ± 127.29} & \multicolumn{1}{|c|}{654.80 ± 285.69} \\ \cline{2-6}
        \multicolumn{1}{|c|}{\multirow{3}{*}{}}  & \multicolumn{1}{|c|}{3 UAVs} & \multicolumn{1}{|c|}{681.80 ± 192.50} & \multicolumn{1}{|c|}{582.00 ± 268.86} & \multicolumn{1}{|c|}{533.60 ± 369.07} & \multicolumn{1}{|c|}{675.8 ± 387.96} \\ \hline
        \multicolumn{1}{|c|}{\multirow{3}{*}{9$\times$9}} & \multicolumn{1}{|c|}{Baseline} & \multicolumn{2}{|c|}{1332.00 ± 804.16} & \multicolumn{2}{|c|}{2264.60 ± 1148.34}\\ \cline{2-6}
        \multicolumn{1}{|c|}{\multirow{3}{*}{}}  & \multicolumn{1}{|c|}{2 UAVs} & \multicolumn{1}{|c|}{980.40 ± 522.45} & \multicolumn{1}{|c|}{1107.60 ± 157.47} & \multicolumn{1}{|c|}{1232.00 ± 573.74} & \multicolumn{1}{|c|}{1087.20 ± 549.05} \\ \cline{2-6}
        \multicolumn{1}{|c|}{\multirow{3}{*}{}}  & \multicolumn{1}{|c|}{3 UAVs} & \multicolumn{1}{|c|}{645.00 ± 203.67} & \multicolumn{1}{|c|}{690.60 ± 308.33} & \multicolumn{1}{|c|}{564.40 ± 210.77} & \multicolumn{1}{|c|}{761.00 ± 297.29} \\ \hline
    \end{tabular}
    \caption{Table with the mean and standard deviation of total actions taken by the swarm of UAVs for each map and for each ANN configuration. Generally, the more UAVs in the swarm, the fewer actions it takes to fly over the entire map.}
    \label{tab:table_results}
\end{table}

Considering the number of UAVs, it can be seen that, indeed, the number of actions required decreases as the number of UAVs increases. Thus, the idea that simultaneity of movements in cooperative groups of UAVs benefits the operations in time and efficiency. This decrement is not linearly dependent on the number of UAVs, since it is strongly influenced by the size of the map and the obstacles. Not all of them have the same obstacles. For example, the difference in actions required for the 8$\times$8 map is several times greater in all cases than for the 7$\times$7 map despite being a map with only 15 more cells. 

In the 5$\times$5 cell map (Figure \ref{fig:flight_environment1}) both models present similar behavior, with the proposed one being slightly faster. The local ANNs are faster than the global ones and present less variance. By having less variance it can be inferred that the paths are as optimal as possible and the model has a robust behavior. This same pattern is true in the 6$\times$6 cell map (Figure \ref{fig:flight_environment2}) cell map. Moreover, in this second map the obstacles are not islands to go around, but form corners that force the UAVs to retrace their steps. Having to retrace their steps is what causes such a large increase in the average movement despite having only 11 more cells, of which 4 are new obstacles.

There is a trend change in the 7$\times$7 cell map (Figure \ref{fig:flight_environment3}). Not only because there are more obstacles and the map is larger, but also because the obstacles do not present horizontal or vertical symmetry. For both models, the cases with 2 UAVs present a very abrupt growth, which may indicate that one UAV disturbs too much the paths of the other. Meanwhile, in the proposed model obtained paths with the lowest mean when used globally with respect to when used locally. These paths have a higher variance, which shows that this solution is not as robust as in local ANNs.

For the 8$\times$8 cell map (Figure \ref{fig:flight_environment4}) there is no longer such an abrupt growth in the means of the actions of the paths. In this case, the proposed model is not the one with the lowest mean of movements for 2 UAVs, but it is the one with the lowest variance, so the computation is more accurate.

Finally, in the 9$\times$9 cell map (Figure \ref{fig:flight_environment5}) both models behave similarly. It should be noted that the results are similar to those of the 8x8 map despite being larger, so it can be understood that the layout of the obstacles is more influential to the size of the map.

Statistical tests were performed at a significance level of $\alpha$ = 0.1. First, a Shapiro-Wilk \cite{razali2011power} test of normality was performed to find out which statistical significance test can be applied. Not all distributions appear not to follow a normal distribution (Table \ref{table_shapiro}). This tends to happen in cases with multiple UAVs rather than a single UAV. It may be due to noise generated by some UAVs in the paths of the others (because they have passed through that cell before, because they coincide in the same cell at the same time, etc.). That is, the path of a UAV is affected by the position of others and by its own paths. Therefore, as a UAV moves, it affects the paths of the others.

\begin{table}[!ht]
    \centering
    \begin{tabular}{|c|c|c|c|c|}
        \hline
        Model  & Configuration            & Number of UAVs          & Map Size & p-value        \\ \hline
\multirow{6}{*}{\cite{puente2022uav}}   & \multirow{4}{*}{Global} & 2 UAVs                  & 6x6      & 0.095   \\ \cline{3-5} 
                          &                         & \multirow{3}{*}{3 UAVs} & 5x5      & 0.019   \\ \cline{4-5} 
                          &                         &                         & 7x7      & 0.026   \\ \cline{4-5} 
                          &                         &                         & 8x8      & 0.017   \\ \cline{2-5} 
                          & \multirow{2}{*}{Local}  & \multirow{2}{*}{3 UAVs} & 6x6      & 0.079   \\ \cline{4-5} 
                          &                         &                         & 7x7      & 0.075   \\ \hline
\multirow{6}{*}{Proposed} & \multirow{3}{*}{Global} & \multirow{2}{*}{2 UAVs} & 7x7      & 0.066   \\ \cline{4-5} 
                          &                         &                         & 8x8      & 0.094   \\ \cline{3-5} 
                          &                         & 3 UAVs                  & 5x5      & 0.011   \\ \cline{2-5} 
                          & \multirow{3}{*}{Local}  & \multirow{2}{*}{2 UAVs} & 5x5      & 0.049   \\ \cline{4-5} 
                          &                         &                         & 8x8      & 0.053   \\ \cline{3-5} 
                          &                         & 3 UAVs                  & 6x6      & 0.057   \\ \hline
    \end{tabular}
    \caption{Table with the p-values of the non-normal distributions resulting from performing the Shapiro-Wilk test \cite{razali2011power}.}
    \label{table_shapiro}
\end{table}
 
Since not all the distributions obtained do not follow a normal distribution, a Kruskal-Wallis significance test \cite{mckight2010kruskal} was used to determine whether they follow significantly different distributions. For this test, a significance level ($alpha$) equal to that used for the normality tests was used.

According to the Kruskal-Wallis test, there are distributions that are significantly different. It is necessary to determine which are significantly different from each other, so a series of Tukey's tests \cite{tukey1949comparing} were performed to find out which are significantly different from each other. The same level of significance was also used for the tests. 

\begin{figure}[htp!]
    \centering
    \includegraphics[scale=0.65]{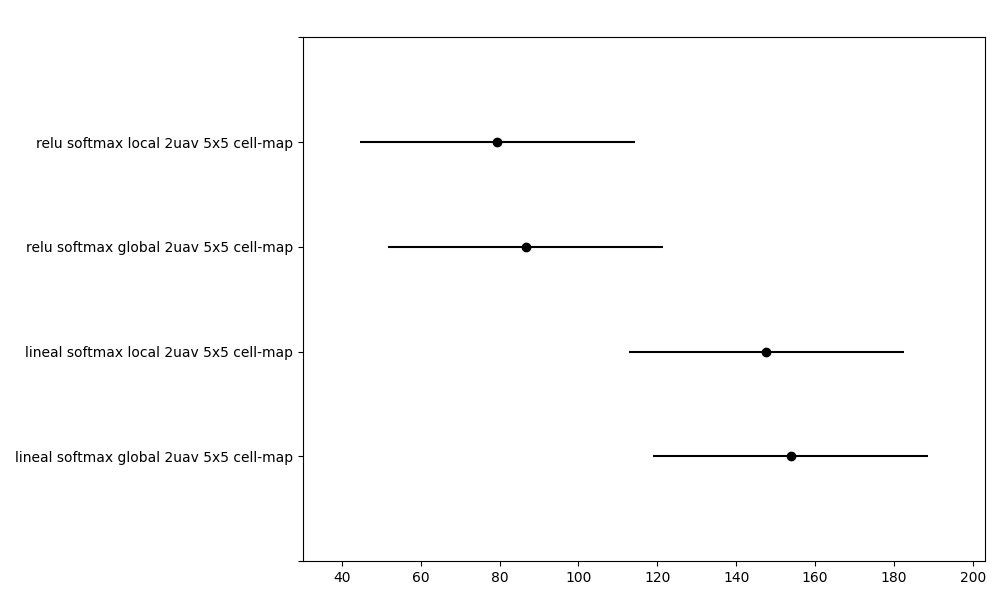}
    \caption{Plot of the universal confidence interval resulting from Tukey's test. The results for the distributions with statistically significant results are displayed. In the $y$ axis, distributions are listed. In the $x$ axis, the average actions taken for the flight paths of each distribution are displayed.}
    \label{fig:tukey}
\end{figure}

The cases in which there has been statistical significance are those resulting from experimenting with 2 UAVs and maps of 5$\times$5. As can be seen in Figure \ref{fig:tukey}, the indicated distributions show differences if the proposed model is compared with the one used for the contrast \cite{puente2022uav}. These show around the half of the average number of actions ($x$ axis) in the proposed model than with the one used to contrast the results. It may be indicative of the models having non-significantly different behavior in all cases except for the indicated cases of 2 UAVs on 5$\times$5 cells maps. Therefore, it is preferable to use the proposed model as it presents shorter or equally sized paths. Although there is no significant difference, in other cases the proposed model also has shorter paths. Such small improvements can also be very advantageous in real environments. That is, by being shorter, the paths allow to save more energy when flying, however little it may be.

Both the proposed model and the one involved in the contrast show no significant differences when comparing their global and local variants for the same model. This indicates that there may not be significant differences from employing one to the other. Therefore, it would be better to use a single global model for all UAVs since it would require fewer computational resources. In other words, if the system uses a single global model, it only has to store the parameters necessary for this one. Otherwise, the parameters of as many models as many UAVs must be stored. This allows the system to be run on less sophisticated and expensive equipment. In addition, using multiple ANNs implies employing control mechanisms such as having different memories for each one. Hence, it is necessary to store more actions and, implement mechanisms that prevent the memory of one ANN from interfering with the memory of another ANN.


\section{Conclusions}\label{conclusions}

In order to control UAV swarms in maps with obstacles, a system based on Q-Learning approaches is proposed in this study. This paper proposes a system based on Reinforcement Learning that makes use of ANNs formed by dense layers. Some of them had already demonstrated the capabilities of these fully connected ANNs on obstacle-free grid-maps \cite{puente2022uav}. The proposed system demonstrates that fully connected ANNs perform satisfactorily on obstacle maps. The use of an ANN with two dense layers allows for the discovery of flight paths that require fewer actions as the number of UAVs in the swarm increases, with the minimal computational expense. As a result, the system's adaptability to various device types is preferred. Instead of employing a single UAV or manually managing UAVs, this enables users to save money, time, and have higher fault tolerance. 


Since it is not necessary to know the spatial relationship of the obstacles with the rest of the environment, it can be understood that the sequence of movements and the position of the UAVs in the swarm is more important. Thus, the actions of a single UAV affect the paths of the others, since it modifies the reward values perceived by the others.

Like prior state-of-the-art works like those cited on Section \ref{background}, the system has been tested on square cell maps with fixed obstacles. It has not been tested with maps of fixed sizes, like the other projects, but rather with different map sizes. A defined set of 5 square cell maps have been used to successfully evaluate the system's capabilities. These maps have fixed obstacles and come in various sizes. The system's capabilities with swarms in various map sizes and with various obstacle configurations are thus demonstrated. Despite the obstacles, it is not necessary to provide the system with information beyond what it already has from the environment. In other words, unlike other state-of-the-art publications, it is not necessary to include targets or other metrics to guide the computation of paths.

The map set has been tested with combinations of 1, 2, and 3 UAVs. The system has had satisfactory results. Solutions have fewer movements the more UAVs there are. In spite of this, it does not show a linear decrease. That is, 2 UAVs do not necessarily perform the task in half the time of 1 UAV. Nevertheless, the effort is reduced. 

The configuration of the ANN with respect to UAVs (global or local) does not seem to have significant differences in the average number of movements required to reach the goals. However, when local ANNs are used, lower standard deviations tend to be encountered. This is indicative of more robust path predictions. It may be because each ANN only specializes in a single UAV, without having to learn something from the others. Therefore, it can be said that it is better to use local ANNs than to have a global ANN for the whole system.

Among the limitations of the system is, mainly, the atomicity of the movements. Performing atomic movements allows for maximizing data capture since the UAV only moves between two adjacent cells. In addition, it is not necessary to invest time and resources in the computation of the smoothing of the path to shorten it and eliminate cycles. In contrast, in some tasks, it may be more optimal to have smoothed path, since it is important to do them in less time to maximize data capture.

Finally, the height of the UAVs is not taken into account. This can influence the calculation of the path since the data collected in a cell at one height does not have to be the same as if it were made at another height. Thus, the reward in these cases could be affected based on the quality of the information captured. UAVs usually fly at altitudes that allow margin in case of disturbances. Moreover, the height with respect to the ground would only have to be varied in case of flying obstacles (such as birds) and would only involve moving a few meters. Despite this, the system has satisfactory results regardless of the flight height.

\section{Future Work}\label{future_work}

This work provides a basis for further investigation on UAV swarms for Path Planning. Especially focused on experiments with small full-connected ANNs in maps with obstacles.

Experiments can be performed with more complex maps than the 2D maps used. For example, experiments with 3D maps, where UAVs can do more types of motions, such as pitch or roll, will be among future improvements. In addition, setting up movements, such as stopping, can reduce the possibility of collisions on overlapping paths.

Better precision in the movement can be achieved at the cost of more complex systems. For example, combining ANNs for different functionalities. Combining multiple ANNs can allow adding capabilities to the flight, such as height or tilt. By having multiple ANNs to determine combined movements (such as climbing while turning), more accurate and faster results can be achieved. Also, using non-discretized maps in future experiments combined with RL techniques such as Double Q-Learning may allow having more precise control of UAVs and, in addition, not limiting the movement to a few positions set in cells. However, having a continuous map implies having a much larger exploration tree, so the computational cost may be increased.

Continuing with greater freedom of movement, future experiments can be focused on how agents learn how to lower hazards during flights through experiments using maps that include fixed and dynamic obstacles implying height estimation. For example, birds, other UAVs, and other objects can be dynamic obstacles.

The most important improvement is to achieve a system that allows a greater variety of movements. For example, these actions can be combinations in different degrees of the above. Even, use the stop as an action. Having more actions and some combined ones makes it more difficult to count the paths, but it can improve the precision of the movements. In this way, the data capture is optimized and the risk of maneuvers is reduced.

\section*{Funding}

This project was supported by the FCT - Foundation for Science and Technology, in the context of the project UIDB/00127/2020, and also POCI 2020, in the context of the Germirrad project - POCI-01-0247-FEDER-072237. Also, the General Directorate of Culture, Education, and University Management of Xunta de Galicia ED431D 2017/16. This work was also funded by the grant for the consolidation and structuring of competitive research units (ED431C 2022/46) from the General Directorate of Culture, Education and University Management of Xunta de Galicia, and the CYTED network (PCI2018\_093284) funded by the Spanish Ministry of Innovation and Science. This project was also supported by the General Directorate of Culture, Education and University Management of Xunta de Galicia "PRACTICUM DIRECT" Ref. IN845D-2020/03.

\section*{Conflict of interest}
The authors declare that they have no conflict of interest.

\section*{Code availability}

Source code and a Docker container are available at:

\url{https://github.com/TheMVS/UAV_SWARMS_RL_FIXED_OBSTACLES_MAPS} 

\url{https://hub.docker.com/repository/docker/themvs/uav_swarms_rl_fixed_obstacles_maps/} 

\bibliographystyle{unsrt}  
\bibliography{references}  

\end{document}